\documentclass[runningheads]{llncs}
\usepackage{graphicx}
\usepackage{array}
\usepackage{hhline}
\usepackage{hyperref}
\usepackage{url}

\usepackage{color}
\usepackage{amsmath}
\usepackage{amsfonts}
\usepackage{nicefrac}

\usepackage{graphicx}
\usepackage{epstopdf}
\usepackage{subfigure}
\usepackage{appendix}

\newcolumntype{L}{>{\centering\arraybackslash}m{5cm}}
\newcolumntype{M}{>{\centering\arraybackslash}m{2.4cm}}
\newcolumntype{D}{>{\centering\arraybackslash}m{1.95cm}}
\newcolumntype{B}{>{\centering\arraybackslash}m{1.8cm}}
\newcolumntype{C}{>{\centering\arraybackslash}m{1.0cm}}

\newcommand{\brac}[1]{\left[#1\right]}

\begin{document}
\title{Scalable Deep Unsupervised Clustering with Concrete GMVAEs}
%
%
\author{Mark Collier \and
Hector Urdiales}
\authorrunning{Collier \& Urdiales}
%
\institute{HubSpot Inc.\, Dublin, Ireland \\
\email{\{mcollier,hector\}@hubspot.com}}
\maketitle              
\begin{abstract}

Discrete random variables are natural components of probabilistic clustering models. A number of VAE variants with discrete latent variables have been developed. Training such methods requires marginalizing over the discrete latent variables, causing training time complexity to be linear in the number clusters. By applying a continuous relaxation to the discrete variables in these methods we can achieve a reduction in the training time complexity to be constant in the number of clusters used. We demonstrate that in practice for one such method, the Gaussian Mixture VAE, the use of a continuous relaxation has no negative effect on the quality of the clustering but provides a substantial reduction in training time, reducing training time on CIFAR-100 with 20 clusters from 47 hours to less than 6 hours.

\end{abstract}

\section{Concrete GMVAE}

Variational Autoencoders (\textbf{VAE}) \cite{kingma2013auto,rezende2014stochastic} are popular latent variable probabilistic unsupervised learning methods, suitable for use with deep neural networks. The standard VAE formulation has a single continuous latent vector in its probabilistic model. However traditional clustering models such as the Gaussian Mixture Model contain a discrete latent variable representing the cluster id. While one can perform clustering using a standard VAE by for example first training a VAE and then performing K-means clustering on the inferred latent variables for each training example, it may be beneficial from a modelling and computational point of view to train end-to-end a VAE capable of clustering via a discrete latent variable. 


The Gaussian Mixture VAE (\textbf{GMVAE}) \cite{dilokthanakul2016deep,ruishugmvae} is one of a number of VAE variants with discrete latent variables which can be used for unsupervised clustering and semi-supervised learning \cite{kingma2014semi}. The GMVAE defines the following generative model of the observed data $\mathbf{x}$:

\begin{align}
p(\mathbf{x}, \mathbf{y}, \mathbf{z}) &= p(\mathbf{y})p(\mathbf{z} \mid \mathbf{y})p(\mathbf{x} \mid \mathbf{z})\\
\mathbf{y} &\sim \text{Cat}(1/K) \\
\mathbf{z} &\sim \mathcal{N}(\mu_z(\mathbf{y}), \sigma_z^2(\mathbf{y}))\\
\mathbf{x} &\sim \mathcal{B}(\mu_x(\mathbf{z})),
\end{align}

Where $\mu_z$, $\sigma_z^2$ and $\mu_x$ are chosen to be deep neural networks. When $\mathbf{x}$ is not distributed Bernoulli a different likelihood model is used. As is standard for VAEs we introduce a factorized inference model $q(\mathbf{y}, \mathbf{z} \mid \mathbf{x}) = q(\mathbf{y} \mid \mathbf{x}) q(\mathbf{z} \mid \mathbf{x}, \mathbf{y})$ for the latents which are also deep networks. The collection of networks are then trained end-to-end to maximize a single sample Monte Carlo estimate of the ELBO $\mathcal{L}$, a lower bound on $\log p(\mathbf{x})$:

\begin{align}
\mathcal{L} &= \mathbb{E}_{q(\mathbf{y}, \mathbf{z} \mid \mathbf{x})}\brac{ \ln \frac{p(\mathbf{y})}{q(\mathbf{y} \mid \mathbf{x})} + \ln \frac{p(\mathbf{z} \mid \mathbf{y})}{q(\mathbf{z} \mid \mathbf{x}, \mathbf{y})} + \ln p(\mathbf{x} \mid \mathbf{y}, \mathbf{z})} \\
&\approx \sum_{\mathbf{y}} q(\mathbf{y} \mid \mathbf{x}) \bigg(\ln \frac{p(\mathbf{y})}{q(\mathbf{y} \mid \mathbf{x})} + \ln \frac{p(\mathbf{z} \mid \mathbf{y})}{q(\mathbf{z} \mid \mathbf{x}, \mathbf{y})} + \ln p(\mathbf{x} \mid \mathbf{y}, \mathbf{z})\bigg)
\end{align}

with $\mathbf{z} \sim q(\mathbf{z} \mid \mathbf{x}, \mathbf{y})$. Thus even the single sample Monte Carlo estimate of the ELBO requires a summation over all possible settings of the cluster label $\mathbf{y}$, making training time linear in the number of clusters used. When one wishes to use a large number of clusters, this linear training time complexity is prohibitive especially when combined with the large datasets and deep neural networks the GMVAE is designed to be applied to.

But the summation over cluster ids is only required because sampling from the categorical distribution $q(\mathbf{y} \mid \mathbf{x})$ is non-differentiable. To avoid the linear scaling of the training time complexity we could use a REINFORCE style gradient estimator \cite{RN55} but such methods tend to provide high variance gradient estimates and corresponding slow practical convergence. We instead propose the \textbf{Concrete GMVAE} by continuously relaxing $q(\mathbf{y} \mid \mathbf{x})$ using the Concrete (also known as Gumbel-Softmax) distribution \cite{maddison2016concrete,jang2016categorical}. As $q(\mathbf{y} \mid \mathbf{x})$ is now a continuous approximation to the discrete Categorical distribution, sampling from $q(\mathbf{y} \mid \mathbf{x})$ is differentiable using the reparameterization trick \cite{kingma2013auto,rezende2014stochastic,maddison2016concrete,jang2016categorical}. It follows that the single sample Monte Carlo estimate of the ELBO can be obtained in a time independent of the number of clusters used:

\begin{equation} \label{eq:naive_elbo}
    \mathcal{L}_{Concrete} \approx \ln \frac{p(\mathbf{y})}{q(\mathbf{y} \mid \mathbf{x})} + \ln \frac{p(\mathbf{z} \mid \mathbf{y})}{q(\mathbf{z} \mid \mathbf{x}, \mathbf{y})} + \ln p(\mathbf{x} \mid \mathbf{y}, \mathbf{z})
\end{equation}

where $\mathbf{y}, \mathbf{z} \sim q(\mathbf{y}, \mathbf{z} \mid \mathbf{x})$, which can be obtained through ancestral sampling. In practice, as has been found when applying VAEs to text \cite{bowman2016generating,yang2017improved}, we found that the direct use of eq.\ (\ref{eq:naive_elbo}) leads to the network ignoring the latent variable $\mathbf{y}$ by reducing the KL divergence $D_{KL}(q(\mathbf{y} \mid \mathbf{x}) || p(\mathbf{y}))$ to zero. We adopt a similar strategy to \cite{bowman2016generating} and introduce a weight $w_t$ on $D_{KL}(q(\mathbf{y} \mid \mathbf{x}) || p(\mathbf{y}))$ which we anneal from zero to one during the course of training. Our MC estimate of the ELBO thus becomes:

\begin{equation}
    \mathcal{L'}_{Concrete} \approx w_t * \ln \frac{p(\mathbf{y})}{q(\mathbf{y} \mid \mathbf{x})} + \ln \frac{p(\mathbf{z} \mid \mathbf{y})}{q(\mathbf{z} \mid \mathbf{x}, \mathbf{y})} + \ln p(\mathbf{x} \mid \mathbf{y}, \mathbf{z})
\end{equation}

We find this modification to be sufficient to encourage use of $\mathbf{y}$.

\section{Experiments}

We test our Concrete GMVAE on the binarized MNIST and CIFAR-100 datasets \cite{lecun2010mnist,krizhevsky2009learning}. For MNIST we set K=10 clusters, one for each class. For CIFAR-100 in order to keep the standard GMVAE training time reasonable we use K=20 clusters, one for each ``super-class'' in the dataset \cite{krizhevsky2009learning}. We anneal the KL weight following: $w_t = \min(1.0, \exp(\frac{-t}{2000}))$. Further experimental details and neural network architectures are given in appendix A. We note that at test time, for all models we fully discretize $\mathbf{y}$, by one-hot encoding the argmax of  $q(\mathbf{y} \mid \mathbf{x})$.

We can see from table \ref{tab:1} that for both datasets there is no significant difference in the test set log-likelihood for the GMVAE and the Concrete GMVAE i.e. our proposed method learns an equally good model of the data to the standard model. Without KL annealing we obsvserved that at convergence $D_{KL}(q(\mathbf{y} \mid \mathbf{x}) || p(\mathbf{y}))$ was close to zero and thus the model failed to learn to cluster the dataset. The computational advantage of using the Concrete GMVAE is clear from the training times, training the Concrete GMVAE is approximately 4X and 8X faster than the standard GMVAE for K=10 and K=20 respectively. As K grows this gap would further increase.

\begin{table}
\caption{Comparison of Concrete GMVAE vs.\ standard GMVAE. Mean and standard deviation of test set log-likelihood and training time (in hours) over 11 training runs are shown. Training was done on an Amazon EC2 p3.2xlarge instance with 1  NVIDIA V100 Tensor Core GPU and 8 virtual CPUs.}\label{tab:1}
\begin{tabular}{|B|C|M|D|M|D|}
\hline
\textbf{Dataset} & \textbf{K} & \multicolumn{2}{c|}{\textbf{GMVAE}} & \multicolumn{2}{c|}{\textbf{Concrete GMVAE}} \\
\hline
- & - & Test $\mathbb{E}_q \brac{\ln p(\mathbf{x} \mid \mathbf{y}, \mathbf{z})}$ & Training time & Test $\mathbb{E}_q \brac{\ln p(\mathbf{x} \mid \mathbf{y}, \mathbf{z})}$ & Training time \\
\hhline{|=|=|=|=|=|=|}
MNIST & 10 & $-46.61 \pm 0.79$ & $15.73 \pm 4.16$ & $-46.96 \pm 0.81$ & $3.55 \pm 0.69$ \\
\hline
CIFAR-100 & 20 & $-1725.47 \pm 2.20$ & $47.24 \pm 8.34$ & $-1725.87 \pm 2.12$ & $5.73 \pm 0.84$ \\
\hline
\end{tabular}
\end{table}

These results demonstrate that the theoretical reduction in training time complexity from linear to constant scaling in K, has in practice reduced training time substantially in a realistic training setup (see appendix A) with early stopping, etc. Despite significant speedup in training time, our introduction of a continuous relaxation of the latent variable $\mathbf{y}$ in a GMVAE has had no significant negative impact on test set log-likelihood. We hope that the introduction of the Concrete GMVAE will enable the application of GMVAEs to large scale problems and problems requiring a large number of clusters where the required training time was previously considered prohibitive.

\bibliographystyle{splncs04}
\bibliography{references}

\appendix \label{appendix:a}
\section{Experimental Details}

For both datasets we train using the Adam optimizer \cite{kingma2014adam} with initial learning rate of $10^{-3}$, for up to 200 epochs, stopping training early if the validation set loss has not improved for 10 consecutive epochs. Each image is passed through a shared encoder convolutional neural network with 20 5x5 filters with stride 1 followed by 40 5x5 filters with stride 1, each layer is followed a 2x2 max pooling layer with a stride of 2 and the RELU activation function and finally a fully-connected layer with 512 units and RELU activation. From this shared encoding of the raw image, we compute $q(\mathbf{y} \mid \mathbf{x})$ with two further fully-connected layers of 512 and 256 units also with RELU activation. Whether sampling $\mathbf{y}$ or marginalizing $\mathbf{y}$ out, $q(\mathbf{z} \mid \mathbf{x}, \mathbf{y})$ is computed by concatenating the output of the shared encoder with the one-hot encoded $\mathbf{y}$ and passing it through two fully-connected layers of 512 and 256 units with RELU activation. $p(\mathbf{z} \mid \mathbf{y})$ is implemented as a single fully-connected layer from the one-hot encoded $\mathbf{y}$. $p(\mathbf{x} \mid \mathbf{y}, \mathbf{z})$ is implemented with a single fully-connected layer followed by two layers of transposed convolutions to those used for the shared encoder. The temperature of the Concrete distribution is set to 0.3 for both datasets throughout training.

\end{document}